\documentclass[letterpaper, 10 pt, conference]{ieeeconf}  % Comment this line out if you need a4paper

\IEEEoverridecommandlockouts                              % This command is only needed if 
\usepackage{amsmath,amsfonts,amssymb}
\usepackage{mathtools}
\usepackage{commath}
\usepackage{pdfpages}
\usepackage{siunitx}
\usepackage{textcomp}
\usepackage{float}
\usepackage{authblk}
\usepackage{url}

\newcommand*{\sprime}{^{\prime}\mkern-1.2mu}
\newcommand*{\dprime}{^{\prime\prime}\mkern-1.2mu}
\newcommand*{\tprime}{^{\prime\prime\prime}\mkern-1.2mu}

\title{\LARGE \bf
Optimal Vehicle Path Planning Using Quadratic Optimization \\ for Baidu Apollo Open Platform
}

\author{Yajia Zhang* \and Hongyi Sun \and Jinyun Zhou \and Jiacheng Pan \and Jiangtao Hu \and Jinghao Miao
\thanks{Yajia Zhang (now at Cruise Automation), Hongyi Sun, Jinyun Zhou, Jiacheng Pan (now at Google), Jiangtao Hu, Jinghao Miao are with Baidu USA LLC,
        {1195 Bordeaux Drive, Sunnyvale, CA, USA} \newline *Corresponding author: Yajia Zhang 
        {\tt\small yajia.zhang@getcruise.com} \newline \textbf{The paper was accepted by Intelligent Vehicle Symposium (IV) 2020}
        }
        }

\begin{document}

\maketitle
\thispagestyle{empty}
\pagestyle{empty}

%%%%%%%%%%%%%%%%%%%%%%%%%%%%%%%%%%%%%%%%%%%%%%%%%%%%%%%%%%%%%%%%%%%%%%%%%%%%%%%%
\begin{abstract}
Path planning is a key component in motion planning for autonomous vehicles. A path specifies the geometrical shape that the vehicle will travel, thus, it is critical to safe and comfortable vehicle motions. For urban driving scenarios, autonomous vehicles need the ability to navigate in cluttered environment, e.g., roads partially blocked by a number of vehicles/obstacles on the sides. How to generate a kinematically feasible and smooth path, that can avoid collision in complex environment, makes path planning a challenging problem. In this paper, we present a novel quadratic programming approach that generates optimal paths with resolution-complete collision avoidance capability.
\end{abstract}

%%%%%%%%%%%%%%%%%%%%%%%%%%%%%%%%%%%%%%%%%%%%%%%%%%%%%%%%%%%%%%%%%%%%%%%%%%%%%%%%
\section{Introduction}
The goal of path planning is to find a function $\boldsymbol{q}$ in configuration space $\mathcal{Q}$ that connects the start configuration $q_s$ and goal configuration $q_g$, where each point in $\boldsymbol{q}$ is in collision-free space $\mathcal{Q}_{free}$. Generally, the function $\boldsymbol{q}$ is represented in the form of a sequence of discretized configurations that are connected by local planners. For autonomous vehicles, especially vehicles with passengers, path planning is not only required to compute a safe path but also equally importantly a comfortable one, i.e., paths with gracefully changed geometrical properties. Nevertheless, the computed path must satisfy the nonholonomic constraint of the vehicle as well. In urban driving environment, it is common that the vehicles need constantly wiggling through narrow passages. All these factors make the path planning for autonomous vehicles a challenge problem.

To tackle this, our approach adopts several strategies. First, we transform the planning frame from map frame to a Frenet frame (see Fig. \ref{fig:frenet-frame-framework}). By doing so, we decouple the path planning problem to two phases: for the first phase, we focus solely on obtaining a smooth driving guide line from map data, which is a prerequisite for accurate Map-Frenet frame conversion; for the second phase, we focus on optimization in the Frenet frame for path generation.

The second strategy is the proposition of \textbf{piecewise-jerk} path formulation. Piecewise-jerk formulation represents a path in a Frenet frame using a sequence of densely discretized points according to the guide line's spatial parameter, where each point contains the lateral distance to the guide line, and its first and second-order derivative w.r.t. the spatial parameter. Between consecutive points, a constant third-order term is used to ``connect" them and maintain second-order continuity of the path. This formulation allows the path to achieve flexibility as it is globally discretized, and smoothness as it is locally second-order differentiable throughout.

The third strategy is we use optimization to compute the path in piecewise-jerk path formulation. The optimization procedure searches a safe and kinematically feasible solution from a safe driving corridor in a Frenet frame, which is computed by considering obstacle occupations, road boundaries, etc.

\begin{figure}
    \centering
    \includegraphics[width=0.48\textwidth]{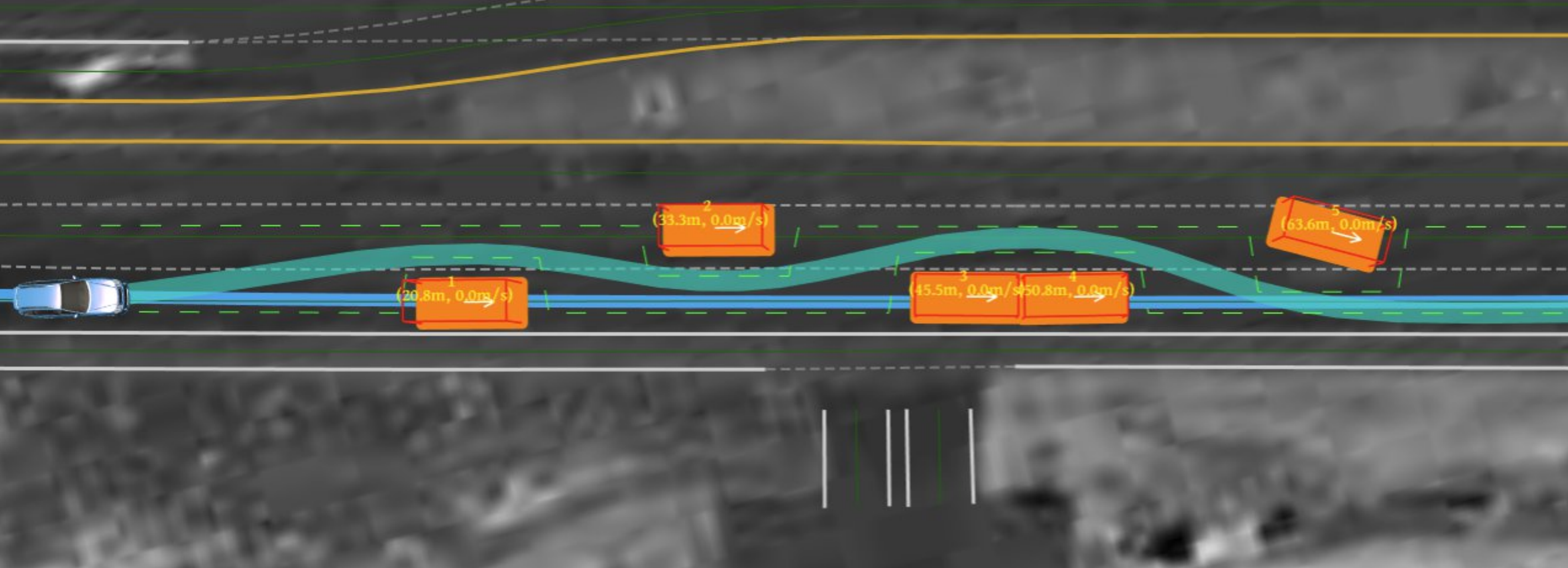}
    \caption{Screenshot of path planning in Apollo Dreamland Simulation environment\cite{baiduapollosimulation}. The driving guide line is shown in blue. The search boundary for path planning (green dashed line) is computed around the guide line by considering the location and geometry of ego vehicle and surrounding obstacles, and road structure. Path optimization procedure computes a path (green strip) with multiple curves to avoid collisions in the cluttered environment.}
    \label{fig:dreamland}
\end{figure}

Our approach is implemented on Baidu Apollo Open platform\cite{baiduapollo} and is tested in both simulation and road tests (see Fig. \ref{fig:dreamland}). The path quality and computation efficiency are widely verified in numerous scenarios. 

\section{Related Work}
Path planning for autonomous driving vehicles has been a vigorous research topic stipulated by DARPA Grand Challenge (2004, 2005) and Urban Challenge (2007). A number of planning algorithms have been developed for tackle the challenge \cite{5940562, miller2008team, urmson2008autonomous, kuwata2009real, thrun2006stanley}.

For path planning, generally the algorithms can be categorized into a couple of classes: randomized planners are intended to solve high-DOF robot motion planning problems as they are good at exploring high dimensional configuration spaces. Some randomized planners, such as Rapidly Exploring Random Tree (RRT)\cite{lavalle2001randomized}, C-PRM\cite{song2001randomized}, can be used for car-like robots with differential constraints, some can even produce high-quality paths given enough computing time \cite{hwan2011anytime}. The problem is they are generally universal planners that cannot effectively exploit the domain knowledge from the road-like structured environment, thus the quality of the paths cannot meet the requirement for comfortable high-speed driving. Therefore, these planners are usually deployed for special scenarios, e.g., parking, rather than general on-road driving.

Discrete search methods, such as \cite{kuwata2009real}, compute a path by concatenating a sequence of pre-computed path segments or maneuvers. The concatenation is done by checking whether the ending configuration of one segment is sufficiently close to the starting configuration of the target segment. These methods generally work well for simple environments. However, the number of required maneuvers needs to grow exponentially in order to solve complex urban driving cases. State lattice search methods\cite{pivtoraiko2005efficient} discretize the state space into lattices. They can be seen as special discrete search methods with transitions from one grid to another implicitly indicating a path segment. The shape of the path heavily depends on the discretization of the lattice which constraints the flexibility of the path.

Optimization-based methods are the most flexible methods. The work in \cite{ziegler2014making} runs a quadratic programming procedure in global/map frame to directly compute a trajectory (path and assigned speed profile at the same time). The trajectory is finely discretized in map frame and these discretized positional attributions are served as optimization variables. The optimization procedure iterates to find a trajectory that minimizes the objective function which combines safety measure and comfort factors. The advantage of optimization methods are they provide direct enforcement of optimality modeling. Furthermore, as the path/trajectory is densely discretized to serve as optimization variables, these methods achieve maximal control of the path/trajectory to deal with complex scenarios.

Besides methods performing path/trajectory planning in map frame, some method transforms the planning problem to a different space to reduce planning complexity. In \cite{werling2010optimal}, trajectory planning is performed in a Frenet frame, which is defined w.r.t a smooth driving guide line. The motion of the vehicle is decoupled to two 1-dimensional motions, longitudinal and lateral motions w.r.t. the guide line. For each 1D planning problem, a set of candidate motions are generated in the form of polynomials. Then, lon. and lat. motions are combined and transformed to Cartesian space for selection. The advantage of planning in Frenet frame is it effectively exploits the road structure and achieves clearer scene-understanding.

Our approach adopts a similar idea in \cite{werling2010optimal} but uses optimization on the 1-dimensional motion planning. This overcomes the problem that polynomial functions are not flexible enough for complex driving scenarios. Our method combines merits of optimization and Frenet frame motion decoupling. Furthermore, autonomous driving systems typically require extremely short planning cycle to account for dynamic changes in the environment. Since we formulate the optimization as a \textbf{quadratic programming} problem, which can be efficiently solved in general, our method has great value of practical usage. To our knowledge, no similar work has been done in the past.

\section{Problem Definition}
\label{sec:problem_definition}

\begin{figure}
    \centering
    \includegraphics[width=0.28\textwidth]{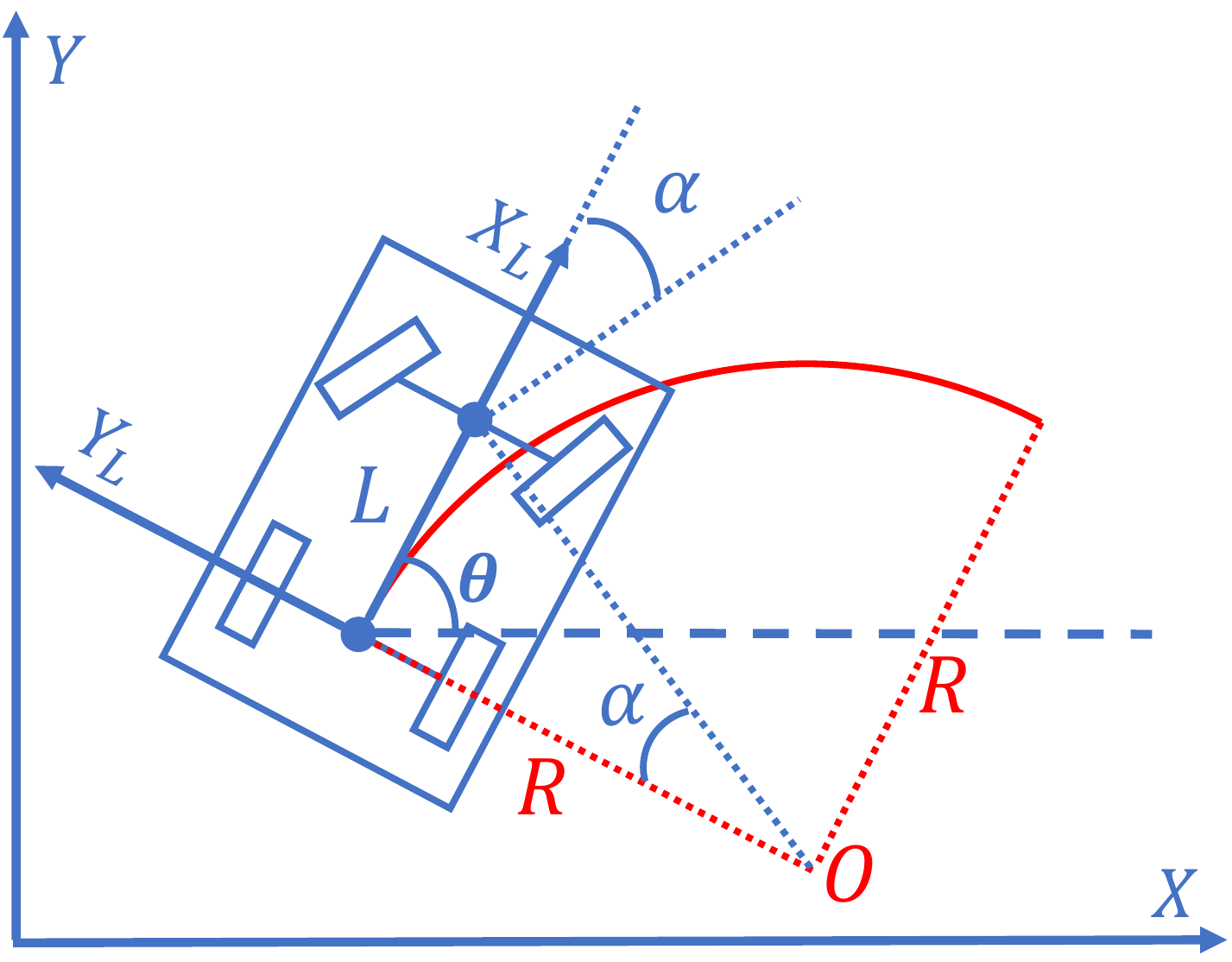}
    \caption{Illustration of vehicle bicycle model and configuration space. The four-wheeled vehicle is simplified to two-wheeled bicycle with one wheel at the center of the front axis and the other at the center of the rear axis. Resulted from the steering angle $\alpha$, the vehicle moves along a circle with radius $R = L/\boldsymbol{\tan}(\alpha)$, where $L$ is the distance between front and rear axis. $\kappa$ in vehicle configuration space is the inverse of $R$, i.e., $\boldsymbol{\tan}(\alpha) / L$, which implicitly represents the steering angle of the vehicle.}
    \label{fig:bicycle-model}
\end{figure}

Generally, the configuration space of a vehicle with differential constraints includes three dimensions, two dimensions $(x, y)$ are used to specify the coordinate of certain reference point of the vehicle, commonly the center of rear axis or center of mass, and one dimension $\theta$ is used to specify the vehicle's heading direction in map frame. Furthermore, we use a bicycle model to model the kinematics of the vehicle (see Fig.\ref{fig:bicycle-model}). The bicycle model assumes the vehicle travel a circle resulted from the steering of its wheels. In our work, besides $(x,y,\theta), $we incorporate one more dimension $\kappa$, which is the curvature of the circle resulted from the steering, into the configuration space to implicitly represent the steering angle. The 4-dimensional configuration space provides more accurate pose description and helps the controller module for better designing feedback controls. 

The goal of path planning is to compute a function $\boldsymbol{q}$ that maps a parameter $p \in [0, 1]$ to a specific configuration $(x, y, \theta, \kappa)$, and satisfies a list of requirements:

\begin{enumerate}
    \item Collision-free: the path cannot lead the vehicle to collide with obstacles. Our overall strategy for trajectory planning divides the collision avoidance to static and dynamic obstacle avoidance, and path planning presented in this paper is targeted to achieve collision avoidance with static obstacles.
    \item Kinematically feasible: the path must be within the kinematic limits of the vehicle so that the vehicle can physically follow.
\end{enumerate}

Nevertheless, comfort is another critical factor to consider, especially for vehicles with human passengers. A comfortable path should have minimal wiggling of steers, reduce unnecessary sharp turns, etc. This factor is seen as a soft requirement of path planning.

\subsection*{Path Planning in a Frenet Frame}
\begin{figure}
    \centering
    \includegraphics[width=0.41\textwidth]{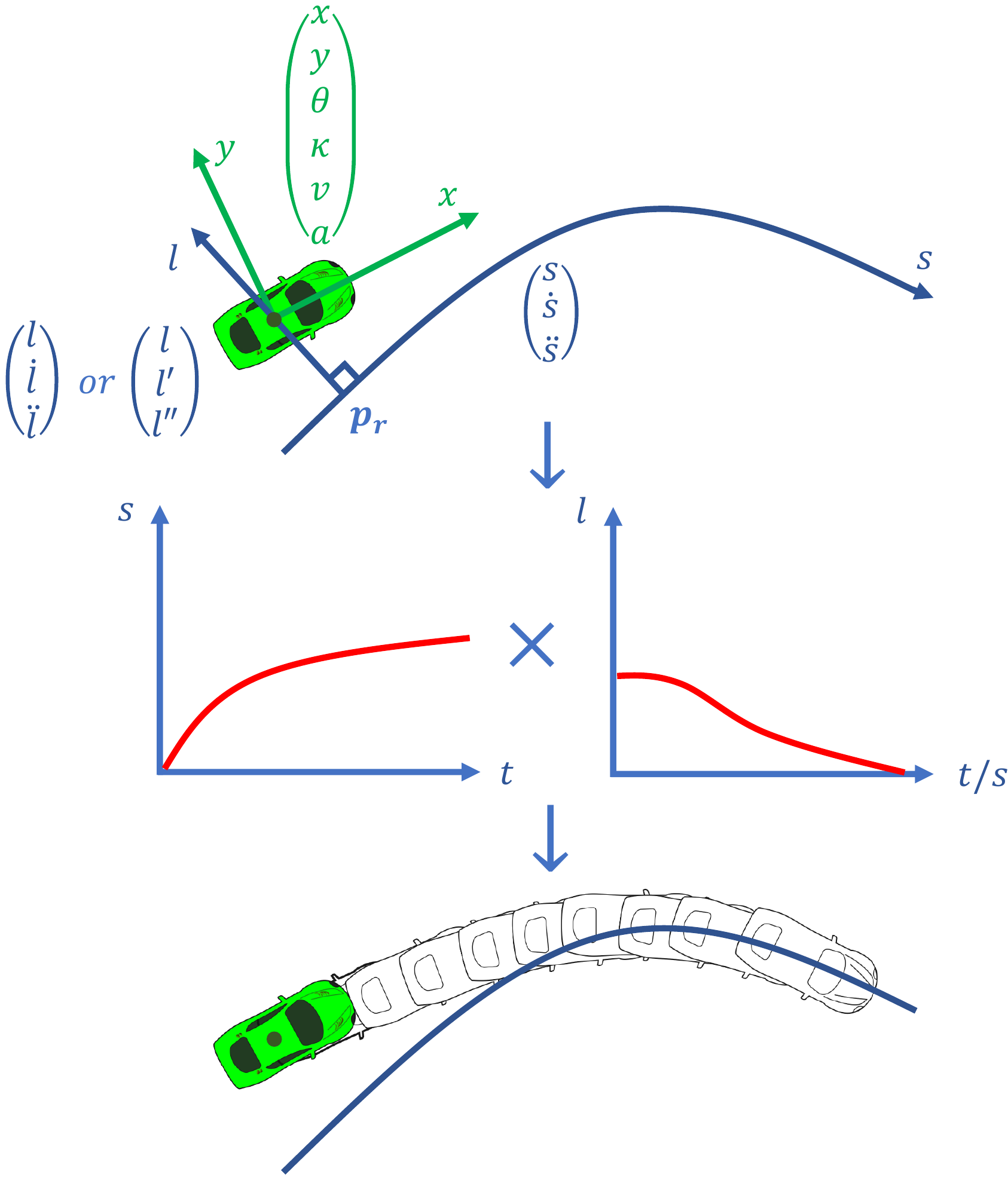}
    \caption{Illustration of vehicle motion planning in a Frenet frame. First, the vehicle state $(x, y, \theta, \kappa, v, a)$, which represents vehicle's position, heading, steering angle, velocity and acceleration in map frame, respectively, is projected on to the driving guide line for motion decoupling. $(s, \dot{s}, \ddot{s})$ represents the vehicle state, i.e., position, velocity and acceleration, along the guide line (i.e., longitudinal state) and $(l, \dot{l}/l\sprime, \ddot{l}/l\dprime)$ represents the vehicle state, i.e., position, velocity and acceleration, perpendicular to the guild line (i.e., lateral state). Then, longitudinal and lateral motions are planned \textbf{independently}. Finally, longitudinal and lateral motions in a Frenet frame are combined and transformed back to map frame.}
    \label{fig:frenet-frame-framework}
\end{figure}

Our method utilizes the concept of Frenet frame to assist on path planning(see Fig. \ref{fig:frenet-frame-framework}). The key idea is to transform the motion planning problem from map frame to Frenet frame, which is defined according to a hypothetical smooth guide line. For vehicle motion planning in a structured environment, the smooth guide line can be the center line of the road that the vehicle targets to follow. %Throughout the paper, we refer the line as driving guide line.

Given a smooth guide line, the vehicle motion in map frame is decoupled into two independent 1-dimensional movements, longitudinal movement that is along the guide line and lateral movement that is orthogonally to the guide line, in a Frenet frame. Thus, a trajectory planning problem is transformed to two lower dimensional and independent planning problems. This framework exploits the task domain that most vehicles are moving along the road, and it is particularly advantageous as it greatly simplifies problem by reducing the dimensionality of planning. 
%Although planning each 1-dimensional movements independently and then combining them together is not fully equivalent to planning directly in map frame, it is an effective way to simplify a high dimensional planning problem with a reasonable trade-off.

Furthermore, instead of representing the lateral movement $\boldsymbol{l}$ as a function of time $t$, another way is to represent it as a function of spatial parameter $s$. Then first and second order derivative $l\sprime = dl/ds$ and $l\dprime = d^2l/ds^2$ represent the first and second order of rate of change of $l$ w.r.t. $s$. This representation essentially defines a geometrical shape, i.e., path, in a Frenet frame. Thus, path planning problem is now finding a function $\boldsymbol{l}(s)$ that satisfies the listed constraints above.

\begin{figure}
    \centering
    \includegraphics[width=0.48\textwidth]{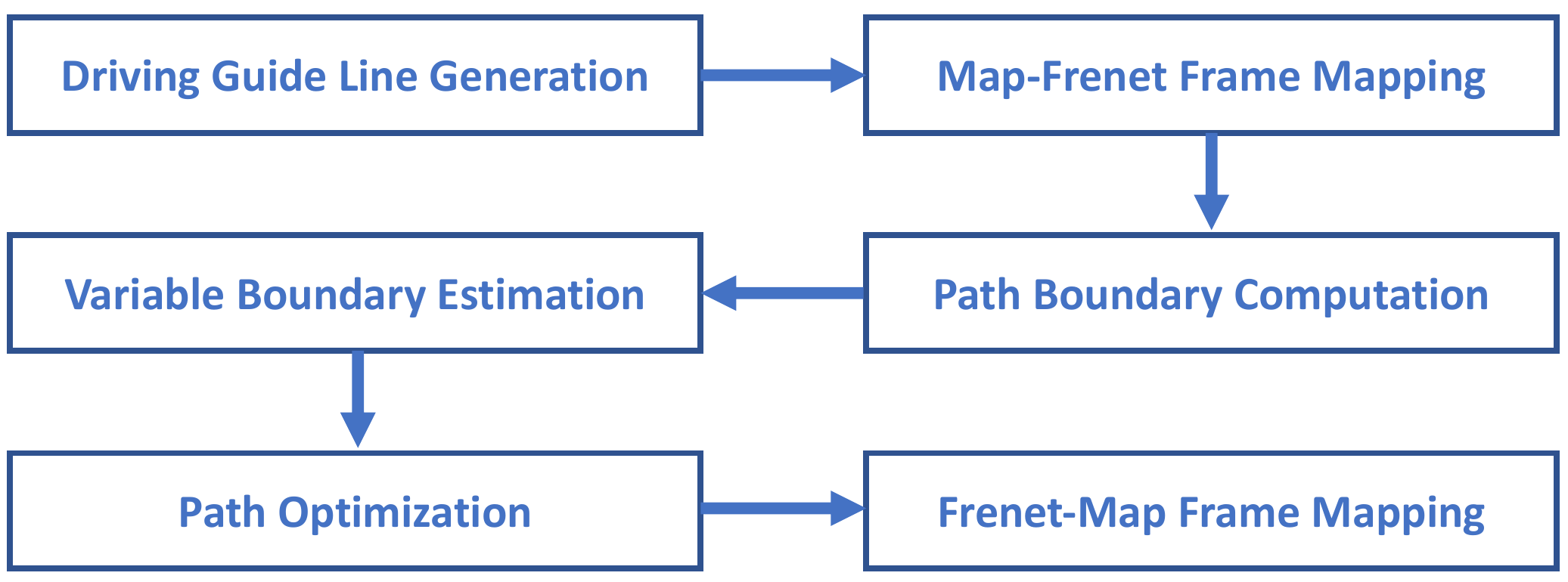}
    \caption{Structure of proposed path optimization approach. Driving Guide Line Generation module (Sec. \ref{sec:guide-line-smoothing}) computes a smooth driving guide line. The smooth driving guide line serves as a ``bridge" to transform the path/vehicle state from map frame to a Frenet frame (Map-Frenet Frame Mapping), and vise versa (Frenet-Map Frame Mapping). Path Boundary Computation module (Sec. \ref{sec:path-boundary-computation}) generates a driving corridor in Frenet frame by considering ego vehicle's position, obstacles' positions and geometries, road structure, traffic rules, etc, for path optimization. Then, Path Optimization module (Sec. \ref{sec:path-optimization}) computes an optimal path within the driving corridor. The kinematic feasibility of the path is ensured by using the variable numerical boundaries derived by Variable Boundary Estimation module (Sub.sec. \ref{subsec:variable-boundary-estimation})}
    \label{fig:data-flow}
\end{figure}

\section{Smooth Driving Guide Line Generation}
\label{sec:guide-line-smoothing}
A guide line is a prerequisite to planning in a Frenet frame. Usually, a guide line is represented as a sequence of positional coordinates from map, i.e, $p_0 = (x_0, y_0), p_1 = (x_1, y_1), \ldots, p_{n-1} = (x_{n-1}, y_{n-1})$, without additional geometrical information, e.g., direction, curvature, etc at any point in the line. Since the guide line is served as the ``bridge" between map and Frenet frame, its smoothness greatly affects the quality of the computed path. According to the Map-Frenet conversion (see \cite{werling2010optimal} for details), to accurately map the path to the curvature level, the guide line must be one order further continuous, i.e., the guide line must be continuous in curvature derivative level.

To fill in the missing geometrical information of the guide line, we develop an optimization-based smoothing algorithm and consider the following aspects in optimization formulation:

\begin{enumerate}
    \item \textbf{Objective} Low and smooth path curvature not only helps reduce instability or overshoot in path tracking by the controller but also enhances the comfort of driving experience. Thus the guide line curvature and its change rate is penalized in optimization. For urban driving scenario targeted by Baidu Apollo Platform, the vehicle is always encouraged to stay in lane. Therefore the guide line is encouraged to be close to the center of the lane.
    % \item \textbf{Line Representation} City road is a mixture of straight and curvy lanes, therefore using piecewise-polynomials to fit the raw guide line is sensitive to input noise, and lead to unexpected oscillations on the smoothed line\cite{ziegler2014making}. In our implementation, we don't represent the guide line as a parametric function but alternatively a sequence of intensively discreted points.
    \item \textbf{Constraints} To account for possible map errors, we allow the input points deviate from their original coordinates to certain extent to achieve possibly higher smoothness while preserving the original shape of the line. Furthermore, to ensure safety, the smoothed guide line must stay within lane boundary. %Feasible range of points $(x, y)$ in optimization are generated by checking lane widths from Apollo High-definition Map, and considering vehicle width and some other road conditions like curb width, etc.
\end{enumerate}

In our past work \cite{zhang2019nonlinear}, we implemented a non-linear optimization algorithm for guide line smoothing. This algorithm is based on the idea introduced in \cite{gulati2013nonlinear}, which models the guide line using a sequence of quintic spiral curves, and computes one sequence that minimizes the spatial length and fluctuations of geometrical properties. This method optimizes the geometrical properties of the line in a direct way. However, the use of spiral curves forms a non-linear relation between the shape of the curve and its positional attributes, thus leads to a difficult and computationally intense non-linear optimization problem. Alternative, in this paper, we present a fast quadratic programming (QP) based algorithm that achieves acceptable empirical results. 

\begin{figure}
    \centering
    \includegraphics[width=0.33\textwidth]{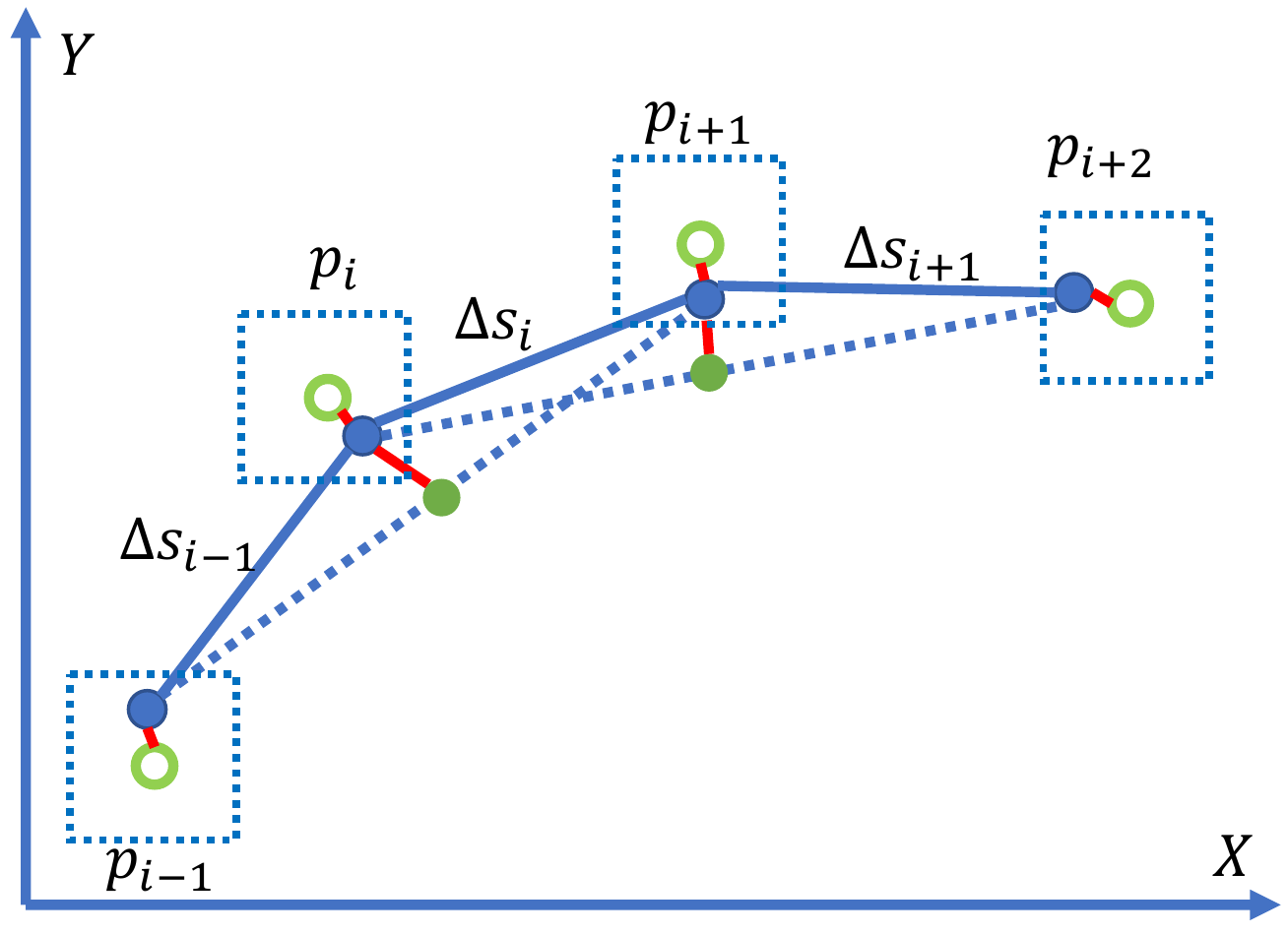}
    \caption{Illustration of guide line smoothing formulation. Given three consecutive points, $p_{i-1}$, $p_i$, $p_{i+1}$, the optimization is set to minimize the distance between $p_i$ and the mid-point of $p_{i-1}$ and $p_{i+1}$, and its deviation to its original input position (green circle in the center of the square box). The positional deviation constraint is simplified to an axis-aligned square box instead of a circle.}
    \label{fig:guide-line-smoothing}
\end{figure}

In formulating the QP problem, the variables are positional variables of the $n$ input points. Given three consecutive points $p_{i - 1}, p_{i}, p_{i + 1}$, we minimize the Euclidean distance from point $p_{i}$ to the middle point of $p_{i - 1}$ and $p_{i + 1}$. Geometrically, the closer $p_{i}$ to the middle point, the straighter these three points, thus the smaller the curvature and its curvature change rate. The points are allowed to deviate from their original positions to certain extent in order to achieve possibly straighter line. The deviations of the points are included as part of the objective function as well (see Fig. \ref{fig:guide-line-smoothing})

% $\mathcal{B}$ is the allowable deviation region for one input point. To simply the constraint, the circle-like region is trimmed and approximated as box-like constraints, i.e., lower and upper bounds for $x_i$ and $y_i$ respectively (see Fig. \ref{fig:guide-line-smoothing}).

% \begin{equation}
%   \label{eq:reference_line}
%   \begin{aligned}
%     & \min\,  
%       \sum_{i=1}^{n - 2} \norm{p_i - 0.5 * (p_{i - 1} + p_{i + 1}) }^2 \\
%     &  \hspace{3.0em} + ~ \sum_{i=0}^{n - 1} \norm{p_i - p_i^{input}}^2
%       \\
%     & \text{subject to:} \\
%     & \hspace{2.0em}
%     p_i \in \mathcal{B}, i = 0, \dots n - 1\\
%   \end{aligned}
% \end{equation}

The output of the guide line smoothing algorithm is a sequence of points with optimized positions. The geometrical information on each discretized points is computed using finite differencing between positions of consecutive points. For any point in between discretized points, its geometrical information is approximated using linear interpolation. 

% \begin{figure}[!b]
%     \centering
%     \includegraphics[width=0.48\textwidth]{figures/path_bound.pdf}
%     \caption{}
%     \label{fig:dreamland}
% \end{figure}

\section{Path Boundary Decision and Computation}
\label{sec:path-boundary-computation}
% In our approach, path optimization is performed in a Frenet frame. Thus, the objective and constraints are formulated based on the generated guide line. In this section, we discuss how we generate feasible search regions for the optimization procedure by processing the diverse inputs, including road data (available lanes at certain section of the road), ego vehicle and static obstacles' location and geometrical information, into uniform expressions that can be described by mathematical inequalities. We achieved this through two steps: lane-utilization decision and path-boundary generation.

In our approach, path optimization is performed in a Frenet frame. Thus, the objective and constraints are formulated based on the generated guide line. In this section, we discuss how we generate feasible search regions for the optimization procedure. It is possible that the complete search region for paths contains a set of geometrical homotopy groups, e.g., passing one static obstacle from either the left or the right side forms two groups. Choosing the best homotopy group is a non-trivial decision making process. In this paper, we propose a heuristic search based decision strategy by taking road data (available lanes at certain section of the road), ego vehicle and static obstacles' locations and geometrical information into account, and formulate mathematical inequalities for the later optimization procedure to use. We achieved this through two steps: lane utilization decision and path boundary generation.

% Advanced techniques such as deep learning might provide better decision by taking more comprehensive information, such as predicted trajectories of dynamic obstacles, road structures from map data, etc, compared to the heuristic based search method proposed above. Our decoupled decision-planning structure is advantageous as decision provides the choice of homotopic search region and planning becomes a purely mathematical problem. 

% In this paper, we propose a heuristic search based decision strategy by processing the diverse inputs, including road data (available lanes at certain section of the road), ego vehicle and static obstacles' location and geometrical information, into uniform expressions that can be described by mathematical inequalities. 

\subsection{Lane Utilization Decision}
In this step, we figure out the lanes on the road for the ego vehicle to use. A naive way is to use the all the lanes available as the drivable area. However, this introduces a few issues:

\begin{enumerate}
    \item  The final generated path may unnecessarily span over a few lanes, which is not only disrespectful to adjacent drivers, but also dangerous.
    \item In cases that the road is narrow (e.g. when in residential area), and there are some other obstacles blocking part of it, we need to temporarily borrow the adjacent or reverse lane to pass.
\end{enumerate}

To tackle the aforementioned problems, we utilize a rule-based decision tree that determines the lanes to use based on traffic rules, ego vehicle state (speed, heading angle, etc.), and blocking obstacle information (obstacle type, position, etc.). This module outputs a set of available lanes for the ego vehicle to use corresponding to the spatial parameter of the guide line. %The lane where ego vehicle stays is prioritized and adjacent lanes are included if necessary.

% It will make sure that, for example, ego vehicle courteously stays in its current lane during normal driving, and when needed, borrow the neighboring lane to generate a detour, etc.

\subsection{Path Boundary Generation}

% Given the decided drivable lanes, the boundary of our path is readily available provided there is no other obstacle, which is rarely the case.
% Therefore, we need to figure out a reasonable area that can carefully and wisely avoid other obstacles.

% Given the decided drivable lanes, the boundary of our path is readily available provided there is no other obstacle, which is rarely the case.
% Therefore, we need to figure out a reasonable area that can carefully and wisely avoid other obstacles.

The goal of this step is to finely process the available lanes from the previous step to specific boundaries by considering vehicle's position and surrounding obstacles.

% The goal of this step is to find a function $\boldsymbol{l}_{B}$ that takes a spatial parameter $s$ as input and outputs a safe boundary, which the vehicle is collision-free with surrounding static obstacles.

To accomplish that, we first discretize the spatial dimension $s$ to a predefined resolution $\Delta s$, along the guide line.
% For each $s_i$, it outputs a safe boundary $(l_{min}, l_{max})$ where the vehicle is collision-free when staying within that bound.
Then starting from the first point $s_0$, we search forwardly. If there are no obstacles at $s_i$, then the available lanes' boundary is directly used as the lower and upper lateral bounds and we proceed to the next point. If static or low-speed obstacles present at $s_i$, we employ a beam-search method in the following way:
based on an estimated ego vehicle's lateral position $l$ at $s_{i-1}$, and the available spacing due to the blocking obstacles, we rank the possible bypassing directions.
Then, we start with the widest feasible one, and search for subsequent $s_{j}$.
If the search along this selection fails in the middle, we back-trace and try other directions.
This procedure is repeated until the searching scope is reached ($160$m in our implementation).
The path boundary, as a result of the above described depth-first search with ranking, will likely be the most reasonable boundary for all the sampled $s_i$.
Note that dynamic or high-speed obstacles are not considered here, as they will be taken care of by the speed planning module.

\begin{figure}
    \centering
    \includegraphics[width=0.42\textwidth]{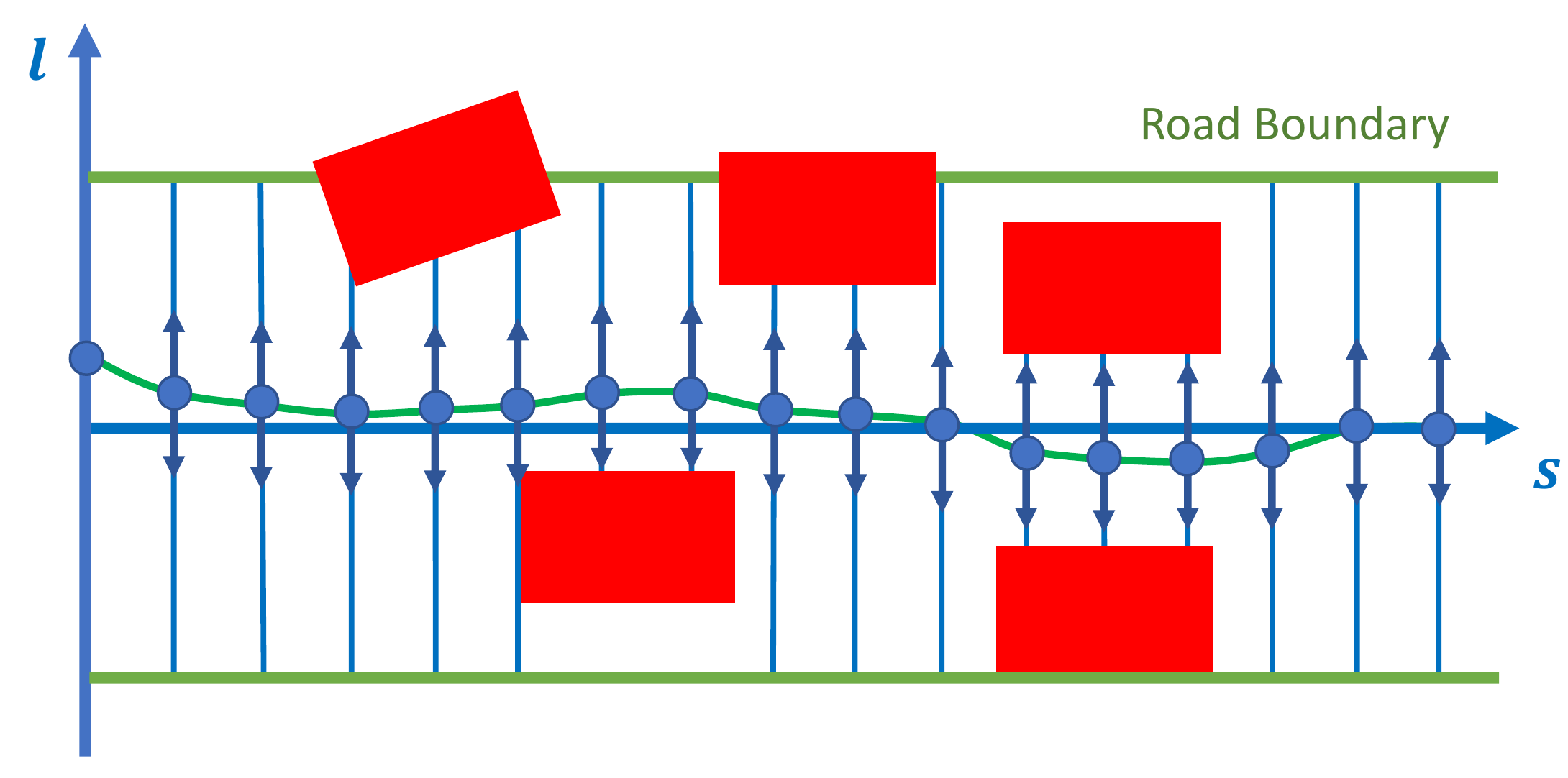}
    \caption{Illustration of path optimization process. Static obstacles are mapped to a Frenet frame given a smooth driving guide line. The spatial axis is discretized to a predefined resolution $\Delta s$. The path boundary computation module computes a feasible boundary function $\boldsymbol{l}_{B}$ for any spatial parameter s. Then the path optimization module optimizes the path by iteratively changing the values for $l$, $l\sprime$, $l\dprime$ at each discretized point.}
    \label{fig:optimization-illustration}
\end{figure}

The output of this step is a function $\boldsymbol{l}_{B}$ that takes a spatial parameter $s$ as input, and outputs a safe boundary $(l_{min}, l_{max})$ for ego vehicle(see Fig.\ref{fig:optimization-illustration}).

% \subsection{Future Direction}
% Choosing the best homotopy search region is a non-trivial decision making process. Advanced techniques such as deep learning might provide better decision by taking more comprehensive information, such as predicted trajectories of dynamic obstacles, road structures from map data, etc, compared to the heuristic based search method proposed above. Our decoupled decision-planning structure is advantageous as decision provides the choice of homotopic search region and planning becomes a purely mathematical problem. 

% Function $\boldsymbol{l}_{B}$'s output is a sequence of boundary points: ${(l_{min}^0, l_{max}^0), (l_{min}^1, l_{max}^1), \ldots,(l_{min}^{n-1}, l_{max}^{n-1})}$ (see Fig.\ref{fig:optimization-illustration}).

\section{Path Optimization}
\label{sec:path-optimization}
In this section, we discuss the detailed optimization formulation in a Frenet frame given the computed path boundary function $\boldsymbol{l}_{B}$.
\subsection{Optimality Modeling}
We consider the following factors in modeling the optimality of a path:

\begin{enumerate}
    \item \textbf{Collision-free} The path must not intersect with obstacles in the environment. 
    \item \textbf{Minimal lateral deviation} If there is no collision risk, the vehicle should stay as close to the center of the lane as possible. 
    \item \textbf{Minimal lateral movement} The lateral movement and its rate of change must be minimized. These two terms implicitly represent how quickly the vehicle moves laterally.
    \item \textbf{(Optional)Maximal obstacle distance} Maximize the distance between the vehicle and the obstacles to allow clearance for the vehicle to pass through safely. This factor turns out can be represented as the distance between the vehicle and the center of its corresponding path boundary. 
\end{enumerate}

The optimization objective for a path $\boldsymbol{l}(s)$ with length $s_{max}$ is defined using the weighted sum of the above factors:
\begin{align*}
    \boldsymbol{f}(\boldsymbol{l}(s)) &= w_l * \int \boldsymbol{l}(s)^2 ds 
                                      + w_{l\sprime} * \int \boldsymbol{l\sprime}(s)^2 ds \\
                                      &+ w_{l\dprime} * \int \boldsymbol{l\dprime}(s)^2 ds 
                                      + w_{l\tprime} * \int \boldsymbol{l\tprime}(s)^2 ds \\
                                      &+ w_{obs} * \int (\boldsymbol{l}(s) - 0.5 * (\boldsymbol{l}_{B}(s)_{min} + \boldsymbol{l}_{B}(s)_{max}))^2 ds
\end{align*}
subject to the safety constraint:
\begin{align*}
    \boldsymbol{l}(s) \in \boldsymbol{l}_{B}(s), \forall s \in [0, s_{max}]
\end{align*}

\subsection{Formulation}
To effectively formulate an optimization problem and evaluate constraint satisfaction in practice, our approach discretizes the path function $\boldsymbol{l}(s)$ according to the spatial parameter $s$ to a certain resolution $\Delta s$, and uses these discretized points to control the shape of the path, and approximately evaluate constraint satisfactions. Same approach has been used in our previously published work \cite{zhang2019nonlinear}. The key idea is to discretize the 1-dimensional function to second-order derivative level, and use constant third-order derivative terms to ``connect" consecutive discretized points. Third order derivative of a positional variable is commonly named jerk. Thus this formulation is named \textbf{piecewise-jerk} method.

\begin{center}
\label{tab:variables}
\begin{tabular}{c c c c c c c c c}
$l_0$        &                         &$l_1$        &                         &$l_2$       &          &$l_{n-2}$        &                         &$l_{n-1}$          \\ 
$l_0\sprime$ &$\xrightarrow{\Delta s}$ &$l_0\sprime$ &$\xrightarrow{\Delta s}$ &$l_2\sprime$&$\ldots$  &$l_{n-2}\sprime$ &$\xrightarrow{\Delta s}$ &$l_{n-1}\sprime$   \\   
$l_0\dprime$ &                         &$l_0\dprime$ &                         &$l_2\dprime$&          &$l_{n-2}\dprime$ &                         &$l_{n-1}\dprime$   \\
\end{tabular}
\end{center}

Table above shows the discretization for path function $\boldsymbol{l}(s)$. $l_i\sprime$ and $l_i\dprime$ are the first- and second-order derivative of variable $l_i$ w.r.t. spatial parameter $s$. $l_i$, $l_i\sprime$ and $l_i\dprime$ at each discretized point are \textbf{variables} in the optimization and they control the shape of the path. 

Between consecutive points, piecewise-jerk method assumes a constant third-order term $l\tprime$ to connect them. The value of the term is obtained from finite differencing the second-order terms:
\begin{equation*}
    l_{i \rightarrow i+1}\tprime = \frac{l_{i+1}\dprime - l_i\dprime}{\Delta s}
\end{equation*}

Note, the third-order term $l\tprime$ is only constant within two consecutive points; among different consecutive points, other values of $l\tprime$ are possible. In order to maintain the continuity of the path, extra equality constraints are introduced between point $i$ and $i+1$:
\begin{align*}
    l_{i+1}\dprime &= l_{i}\dprime + \int_{0}^{\Delta s} l_{i \rightarrow i+1}\tprime ds 
                   = l_{i}\dprime + l_{i \rightarrow i+1}\tprime * \Delta s         \\
    l_{i+1}\sprime &= l_{i}\sprime + \int_{0}^{\Delta s} \boldsymbol{l\dprime}(s) ds 
                   = l_{i}\sprime + l_{i}\dprime * \Delta s + \frac{1}{2} * l_{i \rightarrow i+1}\tprime * \Delta s^2 \\
    l_{i+1}        &= l_{i} + \int_{0}^{\Delta s} \boldsymbol{l\sprime}(s) ds \\
                   &= l_{i} + l_{i}\sprime * \Delta{s} + \frac{1}{2} * l_{i}\dprime * \Delta{s}^2 + \frac{1}{6} * l_{i \rightarrow i + 1}\tprime * \Delta{s}^3 
\end{align*}

This optimization process can be thought as manipulating a stretchable string. We are able to change the shape by applying certain forces and the above equality constraints ensure the string remains one intact piece. The complete piecewise-jerk formulation for path optimization is as follows: find $l_i$, $l_i\sprime$ and $l_i\dprime$, $i \in [0, n-1]$, that minimize
\begin{align*}
    \boldsymbol{\tilde{f}}(\boldsymbol{l}(s)) &= w_l * \sum_{i=0}^{n-1}  l_i^2 
                                      + w_{l\sprime} * \sum_{i=0}^{n-1} l_i\sprime^2  
                                      + w_{l\dprime} * \sum_{i=0}^{n-1} l_i\dprime^2  \\
                                      &+ w_{l\tprime} * \sum_{i=0}^{n-2} \left(\dfrac{l_{i+1}\dprime - l_i\dprime}{\Delta s}\right)^2 \\
                                      &+ w_{obs} * \sum_{i=0}^{n-1} \left( l_i - 0.5 * (l^i_{min} + l^i_{max})\right)^2
\end{align*}

subject to safety and path continuity constraints.

% \newcommand\tr{30}
% \newcommand\trside{10}

% \begin{figure*}[t!]
% \centering
% \begin{tabular}{ccc}
% \label{fig:geometrical-properties}
%       \includegraphics[trim={\trside} {\tr} {\trside} {\tr}, width=0.32\textwidth]{figures/path-properties/theta-s-graph.pdf} & 
%       \includegraphics[
%       trim={\trside} {\tr} {\trside} {\tr}, width=0.32\textwidth]{figures/path-properties/kappa-s-graph.pdf} &
%       \includegraphics[
%       trim={\trside} {\tr} {\trside} {\tr}, width=0.32\textwidth]{figures/path-properties/dkappa-s-graph.pdf} 
% \end{tabular}
% \caption{Geometrical properties of the generated path in Fig. \ref{fig:dreamland}. Vehicle heading $\theta$, instant curvature $\kappa$ and its derivative $\kappa\sprime$ are plotted w.r.t. spatial parameter $s$.}
% \end{figure*}

\begin{figure*}[t!]
    \centering
    \includegraphics[trim={100 10 100 10}, clip, width=0.91\textwidth]{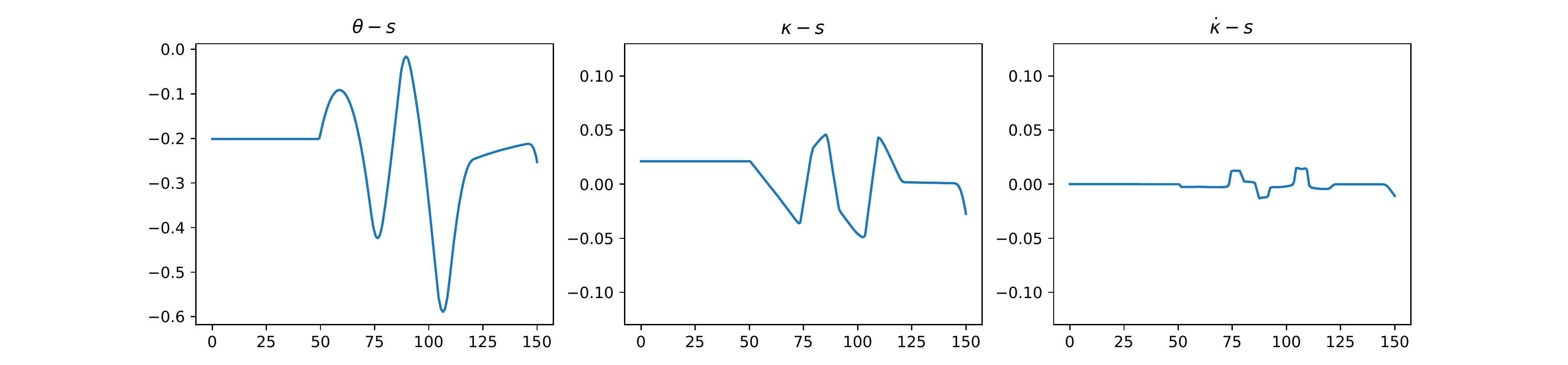}
    \caption{Geometrical properties of the generated path in Fig. \ref{fig:dreamland}. Vehicle heading $\theta$, instant curvature $\kappa$ and its derivative $\kappa\sprime$ are plotted w.r.t. spatial parameter $s$ (unit meter).}
    \label{fig:geometrical-properties}
\end{figure*}

\begin{figure}[t!]
    \includegraphics[width=0.45\textwidth]{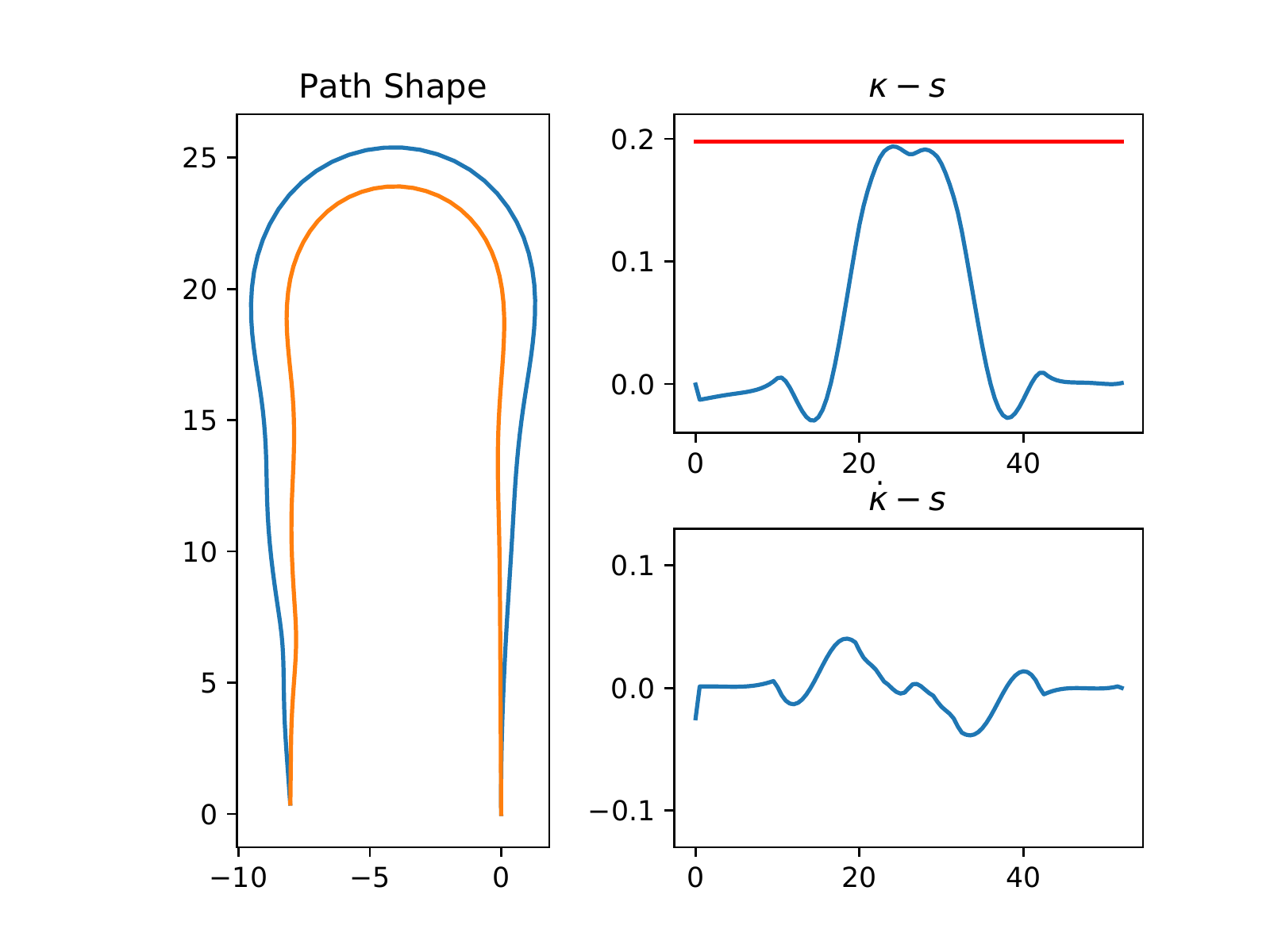}
    \caption{Path planning result of a U-turn case. The guide line is designed to model a sharp turn with maximal curvature exceeding $0.25$. The vehicle is an ordinary sedan with approximately $5.05$m as minimal turning radius. Left: the guide line (orange) and planned path (blue) are shown. Right: curvature and curvature change rate w.r.t. spatial length of the path. The curvature limits for the vehicle is shown as the red line.}
    \label{fig:uturn}
\end{figure}

\subsection{Variable Boundary Estimation for Kinematic Feasibility}
\label{subsec:variable-boundary-estimation}
Beside satisfying geometrical continuity and safety boundary constraints, the computed path must follow the vehicle's kinematic constraint in order to be physically drivable. Since kinematic constraint is defined in map frame, it is difficult to directly enforce this constraint in a Frenet frame due to the complex Map-Frenet frame conversions. In our work, we provide an estimate to the numerical boundaries of variables in the Frenet frame to implicitly enforce the constraint. The most important factor for kinematic feasibility is the curvature of the path. According to the frame conversion equations in \cite{werling2010optimal}, the curvature of one point in path is defined as the following equation:

\begin{equation*}
\label{eq:kappa}
    \kappa = \dfrac{\left(\dfrac{((l\dprime + (\dot{\kappa_r} l + \kappa_r l\sprime) \tan{\Delta\theta}) \cos^2{\Delta\theta}}{1 - \kappa_r  l} + \kappa_r\right) \cos{\Delta\theta} } {1 - \kappa_r  l}
\end{equation*}

where $\kappa_r$ and $\dot{\kappa_r}$ are the curvature and its change rate of the corresponding point $p_r$ on the driving guide line, $\Delta \theta$ is the angle difference between vehicle heading direction and the tangent direction of point $r$.

To simply the complex relation, we make the following assumptions:
\begin{enumerate}
    \item The vehicle is nearly parallel to the driving guide line, i.e., the vehicle's heading angle is assume to be the same as the direction of the guide line at the corresponding point, thus $\Delta\theta = 0$.
    \item The lateral ``acceleration" $l\dprime$ is numerically small (in the order of $10^{-2}$) and is assumed to be $0$.
\end{enumerate}

Based on these, $\kappa$ is approximated as follows:

\begin{equation*}
\label{eq:kappa-approx}
    \kappa \approx \frac{\kappa_r} {1 - \kappa_r * l}
\end{equation*}

Given the kinematic model of the vehicle (see Fig.\ref{fig:bicycle-model}) and vehicle's maximal steer angle $\alpha_{max}$, the maximal curvature for the vehicle can be computed:
\begin{equation*}
    \kappa_{max} = \frac{\boldsymbol{\tan}{(\alpha_{max})}}{L}
\end{equation*}

Thus, we add the linear constraint for $l$ as follows to the optimization procedure for kinematic feasibility:
\begin{equation*}
    % \abs{\dfrac{\kappa_r}{1 - \kappa_r * l}} \leq \frac{\boldsymbol{\tan}{\alpha_{max}}}{L}
    \boldsymbol{\tan}{(\alpha_{max})} * \kappa_r * l - \boldsymbol{\tan}{(\alpha_{max})} + \abs{\kappa_r} * L \leq 0
\end{equation*}

\section{Implementation and Experiments}
The proposed two-step optimization are implemented using Operator Splitting Quadratic Program (OSQP)\cite{osqp}. The source code has been released as part of Baidu Apollo Open Platform.

We use an Intel Xeon $2.2$G HZ computer with $32$GB RAM to run experiments locally and report the results in the paper while comprehensive tests are conducted on Baidu Apollo Dreamland Simulation system with over $200$ synthetic scenarios. 

Guide line smoothing and path optimization are run at each planning cycle with $10$HZ frequency. The total length of guide line to smooth is $300$m with point interval $0.25$m. The average time for computation is $20$ms. Note that guide line smoothing procedure merely depends on map data, thus offline computation of guide line is possible for reducing online computation time. 

For path optimization, the total path length $s_{max} = 150$m with discretization resolution $\Delta s = 0.5$m. The average computation time is $15$ms. The success rate of guide line smoothing and path optimization are $100\%$ on the synthetic scenarios in Baidu Apollo Dreamland. The computed paths are good quality with generally low in curvature fluctuations (see Fig. \ref{fig:geometrical-properties}). 

To test the algorithm's ability to generate kinematically feasible paths, we designed an extremal case in which the maximal curvature of the guide line exceeds the vehicle's physical limits (see Fig. \ref{fig:uturn}). The path takes a larger turn to satisfy the vehicle's kinematic constraints.

% The total time for computing a smooth guide line around 20ms depending on lane curvature conditions using a 2.20GHz computer with 32GB RAM. The experiment result shown in figure ~\ref{fig:ref_line} demonstrates its capability of minimizing curvature while keeping within lane boundary.

% \begin{figure}
%     \centering
%     \includegraphics[width=1.0\linewidth]{figures/ref_line.png}
%     \caption{Screenshot of guide line smoothing in Apollo Dreamland Simulation environment. The red line is the routing segments and the yellow line is the smoothed guide line.}
%     \label{fig:ref_line}
% \end{figure}

% \begin{figure}
%     \centering
%     \includegraphics[width=1.0\linewidth]{figures/ref_line.png}
%     \caption{Screenshot of guide line smoothing in Apollo Dreamland Simulation environment. The red line is the routing segments and the yellow line is the smoothed guide line.}
%     \label{fig:ref_line}
% \end{figure}

% \subsection*{Variable Scaling for Robust Optimization}

\section{Conclusion}
We present a novel optimization-based path planning method for autonomous vehicles. This method decouples path planning into two main stages: in the first stage, a driving guide line smoothing procedure generates a smooth line, which is a prerequisite for planning in a Frenet frame; in the second stage, the path optimizer finds an optimal and kinematically feasible path using piecewise-jerk formulation. Since both stages are formulated as quadratic programming problems, the computation is efficient with average total computation time $40$ms ($20$ms for guide line smoothing and $15$ms for path finding) on a normal PC. The method is released in Baidu Apollo Open Platform and has been deployed on hardware for road test in numerous scenarios.

\addtolength{\textheight}{-12cm}   % This command serves to balance the column lengths

%%%%%%%%%%%%%%%%%%%%%%%%%%%%%%%%%%%%%%%%%%%%%%%%%%%%%%%%%%%%%%%%%%%%%%%%%%%%%%%%

%%%%%%%%%%%%%%%%%%%%%%%%%%%%%%%%%%%%%%%%%%%%%%%%%%%%%%%%%%%%%%%%%%%%%%%%%%%%%%%%

%%%%%%%%%%%%%%%%%%%%%%%%%%%%%%%%%%%%%%%%%%%%%%%%%%%%%%%%%%%%%%%%%%%%%%%%%%%%%%%%
%\section*{ACKNOWLEDGMENT}

\bibliography{reference}{}
\bibliographystyle{plain}

\end{document}